\documentclass[journal]{IEEEtran}

\usepackage{xcolor,soul,framed} %,caption

\colorlet{shadecolor}{yellow}
\usepackage[pdftex]{graphicx}
\graphicspath{{../pdf/}{../jpeg/}}
\DeclareGraphicsExtensions{.pdf,.jpeg,.png}

\usepackage[cmex10]{amsmath}
\usepackage{array}
\usepackage{mdwtab}
\usepackage{eqparbox}
\usepackage{url}
\usepackage{multirow}
\usepackage{amssymb}

\usepackage{array}
\usepackage{graphicx}
\usepackage{hyperref}
\usepackage{bm}

\usepackage{tabularx}
\usepackage{caption}
\usepackage{footnote}
\usepackage{threeparttable}
\usepackage{booktabs}

\begin{document}
\bstctlcite{IEEEexample:BSTcontrol}
    \title{DiffusionVMR: Diffusion Model for Joint Video Moment Retrieval and Highlight Detection}
  %\author{
      %Henghao~Zhao,~\IEEEmembership{}
      % Erez Falkenstein,~\IEEEmembership{Student Member,~IEEE,}
      % and~Zechao~Li,~\IEEEmembership{Fellow,~IEEE}% <-this % stops a space
%}

  %\thanks{Henghao Zhao, Rui Yan, and Zechao Li are with the School of Computer Science and Engineering, Nanjing University of Science and Technology, Nanjing 210094, China. E-mail: \{henghaozhao, ruiyan, zechao.li\}@njust.edu.cn.}% <-this % stops a space
  % \thanks{E-mail: \{henghaozhao, ruiyan, zechao.li\}@njust.edu.cn.}%
  %\thanks{Corresponding author: Zechao Li.}% <-this % stops a space
        % }  
\author{Henghao Zhao,
        Kevin Qinghong Lin,
        Rui Yan
        and Zechao Li
        % <-this % stops a space

% \thanks{Manuscript received xx xx, xxxx; revised xx xx, xxxx.} % <-this % stops a space
\thanks{H. Zhao and Z. Li  are with School of Computer Science and Engineering, Nanjing University of Science and Technology, No. 200 Xiaolingwei Road, Nanjing 210094, China.  E-mail: \{henghaozhao, zechao.li\}@njust.edu.cn.

K. Q. Lin is with the Department of Computer Science and Technology, National University of Singapore, Singapore. E-mail: kevin.qh.lin@gmail.com.

R. Yan is with the Department of Computer Science and Technology, Nanjing University, Nanjing 210023, China. E-mail: ruiyan@nju.edu.cn.

(Corresponding author: Zechao Li)
}}

\markboth{IEEE TRANSACTIONS ON xxxxx, VOL.~xx, NO.~xx, xx~2023
}{Roberg \MakeLowercase{\textit{et al.}}: DiffusionVMR: Diffusion Model for Joint Video Moment Retrieval and Highlight Detection}

% ====================================================================
\maketitle
% === ABSTRACT ====================================================================
% =================================================================================

%%==================================%%
%% abstract %%
%%==================================%%

\begin{abstract}

Video moment retrieval and highlight detection have received attention in the current era of video content proliferation, aiming to localize moments and estimate clip relevances based on user-specific queries. Given that the video content is continuous in time, there is often a lack of clear boundaries between temporal events in a video. This boundary ambiguity makes it challenging for the model to learn text-video clip correspondences, resulting in the subpar performance of existing methods in predicting target segments. To alleviate this problem, we propose to solve the two tasks jointly from the perspective of denoising generation. Moreover, the target boundary can be localized clearly by iterative refinement from coarse to fine. Specifically, a novel framework, DiffusionVMR, is proposed to redefine the two tasks as a unified conditional denoising generation process by combining the diffusion model. During training, Gaussian noise is added to corrupt the ground truth, with noisy candidates produced as input. The model is trained to reverse this noise addition process. In the inference phase, DiffusionVMR initiates directly from Gaussian noise and progressively refines the proposals from the noise to the meaningful output. Notably, the proposed DiffusionVMR inherits the advantages of diffusion models that allow for iteratively refined results during inference, enhancing the boundary transition from coarse to fine. Furthermore, the training and inference of DiffusionVMR are decoupled. An arbitrary setting can be used in DiffusionVMR during inference without consistency with the training phase. Extensive experiments conducted on five widely-used benchmarks (i.e., QVHighlight, Charades-STA, TACoS, YouTubeHighlights and TVSum) across two tasks (moment retrieval and/or highlight detection) demonstrate the effectiveness and flexibility of the proposed DiffusionVMR.

\end{abstract}

% === KEYWORDS ====================================================================
% =================================================================================


% ====================================================================
% ====================================================================
% ====================================================================
% ====================================================================

% \fi
\IEEEpeerreviewmaketitle

% ====================================================================
% ====================================================================
% ====================================================================

% === I. INTRODUCTION =============================================================
% =================================================================================

%%%%%%%%%%%%%%%%%%%%%%
% Introduciton
%%%%%%%%%%%%%%%%%%%%%

\section{Introduction}

\IEEEPARstart{T}{he} rapid development of the Internet has elevated video to the most popular medium in our daily life, favored for its diverse and engaging content~\cite{PAMI_Survey}. However, due to the massive scale, it could be time-consuming and frustrating for individuals to sift through videos manually. Therefore, enhancing search efficiency and content navigability is vital for augmenting user experience on video platforms. These challenges can be tackled by delving into two tasks of video comprehension: video moment retrieval that can automatically identify moments most relevant to user queries. Video highlight detection that can generate concise highlights for a quick overview of video content.

\begin{figure}[!tp]
  \centering
  \includegraphics[width=\linewidth]{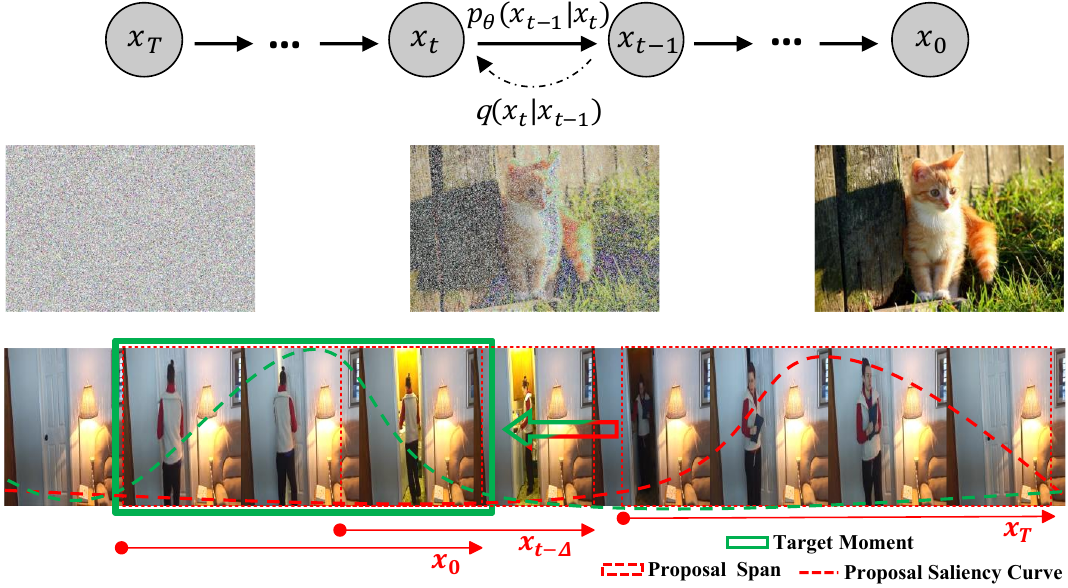}
  \caption{The video moment retrieval and highlight detection can be analogized as an image denoising generation problem: a temporal span(saliency score vector) initialized from noise and progressively refined to cover(highlight)  the target moment.}
  \label{fg1}
  \vspace{-0.5 cm}
\end{figure}

\textbf{Video Moment Retrieval} is defined as localizing consecutive temporal windows in an untrimmed video by giving natural language queries~\cite{PAMI_Survey, twostage2}. The methods designed for this task can be roughly divided into proposal-based and proposal-free paradigms. Proposal-based methods pre-cut the video into various candidate clips, using techniques such as sliding windows~\cite{ twostage2, twostage1, twostage3}, proposal generation networks~\cite{twostage5, twostage6} or multi-scale anchors~\cite{anchor1, anchor2, 2D-TAN}, serving as the support set for retrieval. Although these methods are intuitive and effective, they require additional storage to accommodate numerous proposals generated during the algorithm's runtime. Furthermore, they are are sensitive to the manually designed scale of proposals. In contrast, the proposal-free methods~\cite{proposalfree3, TNN_proposalfree_SDN, TNN_proposalfree_noisylabel} directly predict target moments' start and end indices, offering greater efficiency and flexibility for practical applications. However, they are usually inferior in performance to contemporaneous proposal-based methods. \textbf{Video Highlight Detection} focuses on picking out the key segment with the highest worthiness that best reflects the video gist. Early methods~\cite{lessismore, SL_module} for this task are query-agnostic, treating the task as an unimodal visual scoring problem. Most recently, some methods~\cite{moment-detr} have been introduced to tailor highlights based on specific queries. Although the two tasks are intrinsically linked, the \textbf{Joint Moment Retrieval and Highlight Detection} have not been explicitly studied until recently. Lei et al.~\cite{moment-detr} pioneered this field by establishing the QVHighlights benchmark. They introduced a base model for joint task optimization and demonstrated that the two tasks can mutually reinforce each other. Building on this, UMT~\cite{UMT} incorporates the audio modality to enrich information. QD-DETR~\cite{qddetr} introduces saliency tokens and develops negative pairs for contrastive learning. Overall, these two tasks share a common objective of grounding various scale clips based on user-specific queries, collectively called Video Temporal Grounding by some work~\cite{univtg}.

Considering the sequential nature of video content, it is common to encounter blurred boundaries between temporal events in a video. This boundary ambiguity makes it challenging for the model to learn text-video clip correspondences, leading to inadequate performance when directly predicting target boundaries. To alleviate this problem, we propose jointly tackling two tasks from a denoising perspective and localizing the target boundaries through iterative refinement from coarse to fine. Diffusion models entail a stochastic diffusion process that adds noise to the sample and subsequently trains a model to remove the noise, effectively learning the data's original distribution. This work draws inspiration from diffusion models' success in image generation~\cite{diffusion1, diffusion2, diffusion3}, analogizing the video temporal grounding to a noise-to-image generation task, as shown in Figure~\ref{fg1}. For video moment retrieval, a set of temporal spans sampled from the noise is used as initialization. Subsequently, the location and width of spans are gradually optimized until they can accurately cover the target. In the task of highlight detection, saliency scores for each video clip are randomly generated and progressively refined to precisely indicate their relevance to the query. By pioneering the exploration of the diffusion model for joint moment retrieval and highlight detection, we believe that the model can progressively unveil the relationships between text queries, video content, and their respective event boundaries through multiple iterations. Furthermore, the denoising generation process helps alleviate the confusion caused by boundary ambiguity in the model learning process. As a result, diffusion models show considerable potential for video moment retrieval and highlight detection tasks.

To this end, a novel denoising generation framework is proposed in this work, namely DiffusionVMR. DiffusionVMR includes two branches for video moment retrieval and highlight detection, respectively. In the \textbf{Moment Retrieval branch}, the cross-modal encoder performs video-text cross-modal interaction. It aims to output similarity-aware conditions used as a guide to help the model generate predictions. The moment denoising decoder is the crucial component to perform the denoising process, which contains multiple cascading denoising layers. Each layer takes query-based video representation and noisy spans as input, outputting the new spans updated by the denoised representation. In contrast, the \textbf{Highlight Detection branch} contains only a saliency denoising decoder.  It generates a saliency score vector reflecting the correlation between the query and the video clip. During training, Gaussian noise is added to corrupt the ground truth, obtaining multiple noisy span candidates and a noisy saliency score vector. The model is trained to reverse this noise addition process using denoising decoders. In the inference phase, DiffusionVMR initiates directly from Gaussian noise and progressively refines results from random to real.

Overall, the DiffusionVMR framework offers several advantages. Firstly, our DiffusionVMR borrows from the proposal-free paradigm, which samples initial candidates directly from noise. Secondly, the denoising generation process helps the model learn text-video clip correspondences better. Thirdly, benefiting from the incorporation of the diffusion model, DiffusionVMR can iteratively refine results during the inference process. Fourthly, the training and inference of DiffusionVMR are decoupled, resulting in an arbitrary setting can be used in inference without consistency with the training phase. To evaluate the effectiveness of the proposed framework, the experiments are conducted not only on the joint moment retrieval and highlight detection benchmark (QVHighlights~\cite{moment-detr}), but also on individual task datasets for moment retrieval (Charades-STA~\cite{twostage1}, TACoS~\cite{Tacos}) and highlight detection (YouTubeHighlight~\cite{youtube}, TVSum~\cite{tvsum}). Extensive experiments demonstrate DiffusionVMR's effectiveness, especially achieving a 12\% improvement in average mAP on the QVHigilights dataset compared to the baseline~\cite{moment-detr}. The main contributions of this work are summarized in four folds:

\begin{itemize}

\item{A novel framework for joint video moment retrieval and highlight detection is introduced in this paper, termed DiffusionVMR. It is a pioneering approach to solving the two tasks from the denoising generation perspective.}

\item{In this framework, the moment retrieval task is recast to perform conditional denoising generation of temporal span. The inference spans in DiffusionVMR are initiated from Gaussian noise and progressively refined to produce the final output. The decoupling of training and inference in DiffusionVMR enhances the model's flexibility.}

\item{The video highlight detection task is also reformulated as a conditional denoising generation problem. The saliency score for each video clip is randomly initialized and gradually refined to better reflect the correlation between the query and the video clips.}

\item{Extensive experiments conducted across three settings and five challenging datasets demonstrate the effectiveness of the proposed framework.}

\end{itemize}

% === II. Related Works ========================
% =================================================================================

%%%%%%%%%%%%%%%%%%%%%%
% Related Work
%%%%%%%%%%%%%%%%%%%%%

\vspace{0.2 cm}
\section{Related Work}
This section will review the progress of video moment retrieval, video highlight detection, and diffusion models.

\subsection{Video Moment Retrieval}

Moment retrieval aims to locate specific moments in a video based on a text query. The methods in this task can be divided into two paradigms: proposal-based and proposal-free. Proposal-based methods rely on various proposal generation techniques and rank candidate proposals according to the query. Early methods typically employed strategies such as sliding windows~\cite{twostage2, twostage1, twostage3}, proposal generation networks~\cite{twostage5, twostage6}, or 2D maps~\cite{2D-TAN, TNN_2dmap, TNN_2dmap2} for cutting various proposals. Inspired by some works~\cite{objectdete-anchor1, objectdete-anchor2} in object detection, several methods~\cite{anchor1, anchor2, PAMI_anchorbased} have been proposed to generate proposals by defining multi-scale anchors. Overall, these methods can generate proposals on top of the video representation and maintain them sequentially or hierarchically~\cite{TIP_hierarchy}. The proposal-free paradigm directly estimates the start and end boundaries of target moments without proposal candidates. Most proposal-free methods~\cite{proposalfree3, TNN_proposalfree_SDN, TNN_proposalfree_noisylabel, TNN_proposalfree_actionlo_2, TNN_proposalfree_actionlo_1} model the cross-modal interaction to regress the target boundaries. Some other methods~\cite{reinforcement2, reinforcement3} utilize reinforcement learning to simulate human decision-making processes but face challenges in optimization. Moment-DETR~\cite{moment-detr} is a unique proposal-free method that treats the task as a set prediction problem. It aims to train the decoder to learn a range of moment queries with varying temporal scales, determining whether the relevant scale features in the memory could serve as results. Unlike the above methods, our proposed framework follows the proposal-free paradigm but reformulates the task as a conditional denoising generation problem. The inference candidates in DiffusionVMR are initiated directly from Gaussian noise without prior construction.

\subsection{Highlight Detection}

Highlight Detection focuses on identifying engaging or significant segments within the given video. The task is required to assign a saliency score to each video clip and output the highest-scoring clip as the result. Traditionally, datasets~\cite{Video2gif, tvsum, youtube} in this field are query-agnostic, lacking the capability to tailor highlights according to specific queries. Consequently, numerous studies~\cite{lessismore, SL_module, miniNet} treat this task as an unimodal visual scoring problem, resulting in invariant highlights regardless of the user's query. Recently, Lei et al.~\cite{moment-detr} proposed a benchmark, QVHighlights, for joint moment retrieval and highlight detection tasks. This benchmark allows users to customize highlights for a video based on user-specific queries. They utilized the proposed Moment-DETR to assign the saliency scores. Subsequently, UMT~\cite{UMT} expanded the audio modality for enriched information, while QD-DETR~\cite{qddetr} introduced saliency tokens and developed negative pairs for contrastive learning. Generally, current techniques are ranking-based, where models are trained to assign higher scores to highlight clips via hinge loss, cross-entropy loss, contrastive loss, or reinforcement approaches. In contrast, our framework takes a generative perspective to tackle this task and demonstrates that employing a diffusion training strategy yields improved performance.

\subsection{Diffusion Models}

The diffusion models~\cite{diffusion1, diffusion2, diffusion3} are a class of neural networks inspired by thermodynamic stochastic diffusion processes, which effectively learn original data distributions by reversing the noise addition process. It has shown promise in various generative tasks, such as image generation~\cite{diffusionimage1, TNN_diffusion_image}, audio generation~\cite{diffusionaudio1, diffusionaudio2}, and natural language generation~\cite{diffusionlanguage1, diffusionlanguage2}. Due to its powerful modelling capacity, the diffusion model has also been explored in some discriminative visual understanding tasks, such as segmentation~\cite{diffusionsgmtation1, diffusionsgmtation2}, detection~\cite{DiffusionDet, diffusionvad}, retrieval~\cite{diffusionret} and visual grounding~\cite{diffusiongrounding1}.

Among them, DiffsionDet~\cite{DiffusionDet} is a pioneering approach in applying diffusion model to discriminative tasks, explicitly focusing on object detection. DiffVAD~\cite{diffusionvad} is dedicated to unsupervised video anomaly detection tasks. It leverages the reconstruction capability of the diffusion model to identify anomalies indicated by high reconstruction errors. DiffPose~\cite{DiffusionPose} is used for video-based human pose estimation and treats the task as a conditional heatmap generation problem. Inspired by DiffusionDet, DiffTAD~\cite{DiffusionTAD} and DiffAct~\cite{DiffusionACT} explore the application of diffusion models for temporal action detection and temporal action segmentation, respectively. However, discussions regarding video moment retrieval or highlight detection tasks are scarce. Moreover, this study is centred on multimodal video-text tasks. The joint tasks are reformulated as conditional denoising generation processes, yielding diverse results under distinct query conditions. In contrast, the previous methods do not incorporate any conditionality in their execution.

Most recently, we noticed that MomentDiff~\cite{Momentdiff} shares similar motivations with this paper but diverges in implementation details. Crucially, DiffusionVMR redefines the highlight detection task by incorporating a diffusion model. As a contemporaneous work, DiffusionVMR exhibits a substantial performance advantage, owing to our unique implementation and the synergistic effect of combining two tasks. In summary, this work introduces a new framework that employs diffusion models to integrate denoising learning into joint video moment retrieval and highlight detection tasks. This framework avoids reliance on human priors or hand-crafted components, instead directly sampling candidates from noise, and progressively refining them from the noise to meaningful output.

\begin{figure*}[!ht]
  \centering
  \includegraphics[width=1\linewidth]{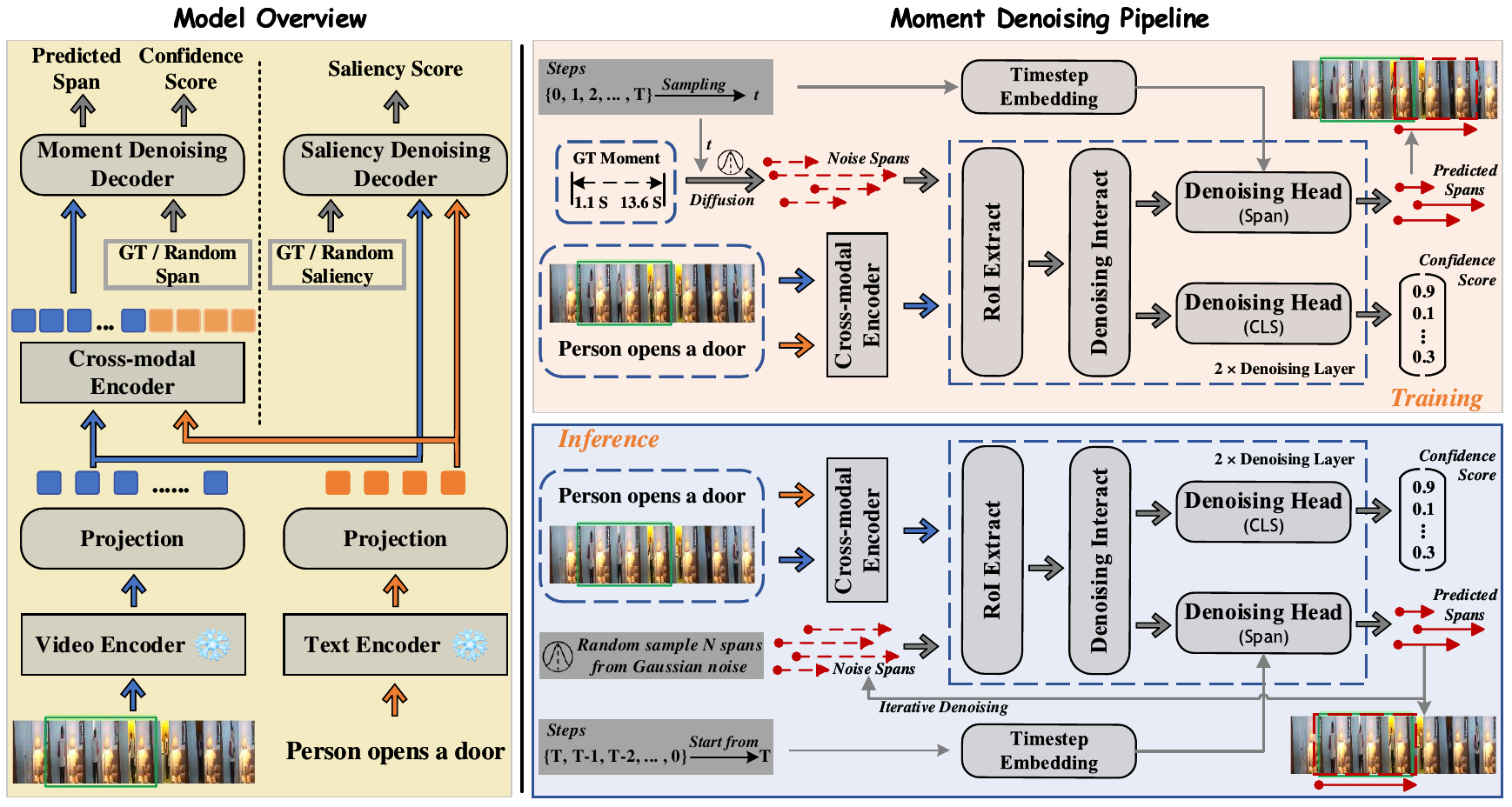}
  \caption{\textbf{Left:} Overview of the proposed DiffusionVMR for joint video moment retrieval and highlight detection tasks. \textbf{Right:} The pipeline of the training and inference of video moment retrieval. The moment retrieval branch comprises a cross-modal encoder and a moment denoising decoder. In the training phase, the GT moment undergoes $t$-step diffusion processes and then is provided to the denoising decoder as the noisy span. Each layer in the denoising decoder takes video representation and noisy span as input, outputting the new span updated by the denoised representation. In inference, the noisy spans are directly sampled from Gaussian noise, and the final output from the decoder is used as the result.}
  \label{fg2}
  \vspace{-0.2 cm}
\end{figure*}

% === III. Methods =======================================
% =================================================================================

%%%%%%%%%%%%%%%%%%%%%%
% Methods
%%%%%%%%%%%%%%%%%%%%%

\vspace{0.2 cm}
\section{Approach}

The diffusion models~\cite{diffusion1, diffusion2, diffusion3} have recently revolutionized the framework for various discriminative tasks~\cite{diffusionsgmtation1, DiffusionDet, diffusiongrounding1} in computer vision, demonstrating impressive performance improvements. This section introduces a framework that integrates the diffusion models to redefine video moment retrieval and highlight detection tasks as unified conditional denoising generation problems. This framework aims to progressively uncover the relationship between text queries, video content, and their corresponding event boundaries with several iterative refinements. Concurrently, the denoising process alleviates the confusion caused by boundary ambiguity during the model's learning phase. 

Given a video-text pair as the input, the video $\bm{V}=[\bm{v}_0, \bm{v}_1, …, \bm{v}_{N_v}]$ comprises $N_v$ clips and the input text query $\bm{Q}=[\bm{q}_0, \bm{q}_1, …, \bm{q}_{N_q}]$ is consistent with $N_q$ word tokens. The objective of joint video moment retrieval and highlight detection is to identify one or multiple pairs of timestamps $(\tau_\mathrm{s}$, $\tau_\mathrm{e})$, while simultaneously predicting a saliency score vector $\bm{l} \in \mathbb{R}^{N_v \times 1} $. Here, $0 \leq \tau_\mathrm{s} \leq \tau_\mathrm{e} \leq N_v$. The model must ensure that video clips within the paired timestamps are semantically related to the text query and assigned a higher saliency score. The moment, defined by paired timestamps, can also be represented as a temporal span $\bm{s}=[s_\mathrm{c}, s_\mathrm{w}]$, where $s_\mathrm{c}=(\tau_\mathrm{e} + \tau_\mathrm{s})/2$, $s_\mathrm{w}=(\tau_\mathrm{e} - \tau_\mathrm{s})$. This section begins with a short preliminary, followed by a succinct overview of the proposed approach. The proposed moment retrieval branch is introduced in Section C and the highlight detection branch is detailed in Section D. Finally, the loss function utilized in this model is presented in Section E.

\subsection{Preliminary}

A concise review of the diffusion models is provided in this section. The forward process of diffusion models follows a Markovian chain. A sequence of Gaussian noise is iteratively added to the ground truth, progressively destroying the original data distribution. The forward process can be represented as,

\begin{equation}
    \bm{x}_t:=\sqrt{\alpha_t}\bm{x}_{t-1}+\sqrt{1-\alpha_t}\epsilon ,
\end{equation}

\noindent where $\bm{x}_{t}$ denotes the noisy sample at diffusion step $t$. $\epsilon$ is the Gaussian noise. $\alpha_{t}=1-\beta_{t}$ is a trade-off parameter determining the noise addition rate. $\beta_{t}$ is a pre-defined noise variance schedule. By following this process, the data distribution for the next diffusion step is derived by adding noise to the previous distribution, forming a Markov chain. As the number of diffusion steps increases, the data distribution gradually worsens and ultimately converges to a Gaussian noise. Actually, $\bm{x}_t$ can be directly derived from $\bm{x}_0$, bypassing recursive steps~\cite{diffusion2},

\begin{equation}
\begin{aligned}
    q(\bm{x}_t \vert \bm{x}_0)&:=\mathcal{N}(\bm{x}_t;\sqrt{\bar{\alpha}_t}\bm{x}_0,(1-\bar{\alpha}_t)\mathbf{I}) \\
              &:=\sqrt{\bar{\alpha}_t}\bm{x}_0+\sqrt{1-\bar{\alpha}_t}\epsilon ,
\end{aligned}
\label{diffusion_process_EQ}
\end{equation}

\noindent where $\bar{\alpha}_t:=\prod_{i=0}^t{\alpha_i}$. $\epsilon \sim \mathcal{N}(\mathbf{0}, \mathbf{I})$ is sampled from a Gaussian distribution.

The reverse process entails progressively mapping the noisy sample $\bm{\hat{x}}_t$ back to the original $\bm{\hat{x}}_0$, forming a sequence $[ \bm{\hat{x}}_t, \bm{\hat{x}}_{t-1}, \cdots, \bm{\hat{x}}_0]$. The procedure can be written as,
\vspace{-1 em}

\begin{equation}
p_\theta(\bm{\hat{x}}_{t-1} \vert \bm{\hat{x}}_t) := \mathcal{N}(\bm{\hat{x}}_{t-1}; \boldsymbol{\mu}_\theta(\bm{\hat{x}}_t, t), \boldsymbol{\Sigma}_\theta(\bm{\hat{x}}_t, t)).
\end{equation}

\noindent Instead of directly parameterizing $ \boldsymbol{\mu}_\theta(\bm{\hat{x}}_t, t) $, the model directly estimates $\bm{\hat{x}}_0$ from $\bm{\hat{x}}_t$ using a neural network $f_\theta(\bm{x}_t,t)$. The iterative denoising process can be redefined as,
% \vspace{-0.5 em}

\begin{equation}
q(\bm{\hat{x}}_{t-1} \vert \bm{\hat{x}}_t, \bm{\hat{x}}_0) := \mathcal{N}(\bm{\hat{x}}_{t-1}; \tilde{\boldsymbol{\mu}}(\bm{\hat{x}}_t, \bm{\hat{x}}_0), \tilde{\beta}_t \mathbf{I}) .
\label{Diff_infence_diffusion}
\end{equation}

\noindent The neural network $f_\theta(\bm{x_t},t)$ can be optimized by minimizing the training objective with MSE loss,
\vspace{-0.2 em}

\begin{equation}
L_\mathrm{train}= 1/2  \Vert f_\theta(\bm{x}_t,t)-\bm{x}_0 \Vert^2.
\end{equation}

\subsection{Model Overview}

The primary objective of DiffusionVMR is to learn the conditional distribution of target video clips through denoising learning. As delineated in Figure~\ref{fg2}, the model contains a moment retrieval branch and a highlight detection branch. In the moment retrieval branch, the cross-modal encoder is used to construct the denoising generation condition, i.e. query-based video representation. The moment denoising decoder comprises multiple cascading denoising layers that restore the noisy spans to the target. Each layer takes the query-based video representation and noisy spans as input, outputting the updated spans refined by denoised representations. The highlight detection branch incorporates a saliency denoising decoder. It processes the video representation, the text representation, and a noisy saliency score vector, with the ultimate objective of generating a saliency score for each video clip.

An off-the-shelf backbone is used to extract representations from video and text queries. More specifically, a sequence of clip embedding $\bm{E}_\mathrm{V}=[\bm{e}_{0}, \bm{e}_{1}, …, \bm{e}_{{N_v}}] \in {{\mathbb{R}} ^ {N_v \times D_v}}$ can be obtained for $N_v$ video clips. The embedding of the text query is represented as $\bm{E}_\mathrm{Q} \in {{\mathbb{R}} ^ {N_q \times D_q}}$. To enable interaction between the video and text modalities, a MLP is used to project the video and text features into a common space with dimension $D$. DiffusionVMR implements a diffusion process $[ \bm{x}_0, \bm{x}_1, \ldots, \bm{x}_t]$ from the ground truth $\bm{x}_0$ to noisy proposal $\bm{x}_t$, where $t \in [0, 1, \ldots , T]$. As the number of diffusion step $t$ increases, $\bm{x}_t$ approaches pure noise. The model is trained to learn the conditional distribution of target clips from the denoising process $\bm{x}_t $ to $\hat {\bm{x}}_0$.

\smallskip
\noindent \textbf{Noisy Proposal Generation.} Before the denoising learning phase commences, it is essential to undertake the diffusion process. This process involves generating noisy proposals from the ground truth. Specifically, it requires preparing noisy proposal spans for the moment retrieval branch and noisy saliency score vectors for the highlight detection branch.

First of all, a diffusion step $t \in [0, 1, \ldots , T]$ is randomly selected to determine the forward diffusion steps. In the moment retrieval task, due to the number of temporal spans varies across videos, random temporal spans are introduced to pad the ground truth to a fixed quantity. The padded ground truth $\bm{x}^{Mo}_0 \in \mathbb{R}^{N \times 2}$ will be normalized. Subsequently, Gaussian noise is added to $\bm{x}^{Mo}_0$ in accordance with Eq.\ref{diffusion_process_EQ} and the chosen diffusion step $t$. The process for the highlight detection task is analogous. The saliency score vector $\bm{x}^{Sa}_0 \in \mathbb{R}^{N_v \times 1}$ is progressively altered according to Eq.\ref{diffusion_process_EQ} and the diffusion step $t$, yielding the noisy saliency score vector $\bm{x}^{Sa}_t$. Notably, for datasets with explicit saliency labels, such as QVHighlight~\cite{moment-detr}, the average saliency score for each clip is utilized instead of one-hot labelling.

\smallskip
\noindent \textbf{Training pipeline.} After the creation of noisy proposal spans and noisy saliency score vector, they are fed into the moment denoising decoder $g_{\varphi}$ and the saliency denoising decoder $h_{\psi}$ for denoising learning, respectively. The moment denoising process can be written as,

\vspace{-.4 em}

\begin{equation}
    \hat{\bm{x}}^{\mathrm{Mo}}_0, \hat{\bm{p}}_c=g_{\varphi}(\bm{x}^{\mathrm{Mo}}_t, t, \bm{M}_\mathrm{V}),
\end{equation}

\noindent where the output $\hat{\bm{x}}^{\mathrm{Mo}}_0 \in {{\mathbb{R}} ^ {N_s \times 2}}$ indicates the predicted spans. $\hat{\bm{p}}_c \in {{\mathbb{R}} ^ {N_s \times 2}}$ are the predicted confidence scores. 

Following~\cite{moment-detr}, a foreground label is assigned if there is a match with the ground truth, and a background label is assigned otherwise. $\bm{M}_\mathrm{V}$ is the query-base video representation obtained from the cross-modal encoder Eq.\ref{encoder_EQ}. The diffusion step $t$ is used to inform the model about the noise intensities. Given that the predictions and the ground truth are not always one-to-one matchings, bipartite matching is used to calculate the loss. Due to the instability of bipartite matching in the early stage of training, it is recommended to gradually increase the number of proposals to enhance convergence speed.

The saliency denoising process can be represented as,

\begin{equation}
    \hat{\bm{x}}^{\mathrm{Sa}}_0 =h_{\psi}(\bm{x}^{\mathrm{Sa}}_t, t, \bm{E}_\mathrm{V}, \bm{E}_\mathrm{Q}),
\end{equation}

\noindent where the output $\hat{\bm{x}}^{\mathrm{Sa}}_0 \in {{\mathbb{R}} ^ {N_v \times 1}}$ is the predicted saliency score vector. $\bm{E}_\mathrm{V}$ is a sequence of video clips embedding and $\bm{E}_\mathrm{Q}$ is the word embedding.

\smallskip
\noindent \textbf{Inference.} Upon completing the training, the DiffusionVMR can reconstruct the moment localization and clip saliency score from arbitrary noise levels. Therefore, proposal spans and vectors are sampled from pure noise as $\hat {\bm{x}}^{\mathrm{Mo}}_T$ and $\hat {\bm{x}}^{\mathrm{Sa}}_T$, which are used to perform the denoising process $[ \hat {\bm{x}}_T, \hat {\bm{x}}_{T-1}, \cdots,\hat {\bm{x}}_0]$. The estimated $\hat {\bm{x}}_0$ can approximate the underlying ground-truth distribution and regard it as the final result. Skipping some sampling steps can accelerate the denoising process~\cite{diffusion2}, allowing for variable numbers of sampling steps during inference. In addition, the number of initial proposal spans during inference does not need to be the same as in training.

\subsection{Moment Retrieval Branch}

\noindent \textbf{Cross-modal Encoder.} The cross-modal encoder aims to produce a query-based video representation that serves as a guide to help the model generate predictions. In detail, video and text features are concatenated to feed into the cross-modal encoder. The encoder $f_{\theta}$ comprises a stack of six-layer Transformers with the multi-head self-attention and a feed-forward network. It models the sequence context and enhances cross-modal interaction. The output of the cross-modal encoder is called Memory, following the convention.

\vspace{-.3em}
 
\begin{equation}
    {[\bm{M}_\mathrm{V},\bm{M}_\mathrm{Q}]}=f_{\theta} {[\bm{E}_\mathrm{V},\bm{E}_\mathrm{Q}]}
    \label{encoder_EQ}
\end{equation}

\noindent where $\bm{M}_\mathrm{V} \in {\mathbb{R} ^ {N_v \times D}} $ and $ \bm{M}_\mathrm{Q}\in {\mathbb{R} ^ {N_q \times D}}$ are the video memory tokens and text memory tokens, respectively. After multiple layers interactions, the encoder can produce video representations with contextual semantic contents and cross-modal similarity relations. Therefore, the representation $M_v$ can be used as a guide for generating predictions.

\smallskip
\noindent \textbf{Moment Denoising Decoder.} The moment denoising decoder aims to restore noisy spans to their original locations progressively. It comprises two cascading denoising layers, each consisting of a RoI feature extraction module, a feature interaction module, and dual-target denoising heads, as depicted in Figure~\ref{fg2}. Specifically, the denoising layer takes a set of noisy proposal spans as input and crops the corresponding clip features from the video memory $\bm{M}_\mathrm{V}$. Then, these features are sent to the interaction module. The final output spans are updated by the denoised representation via the denoising head.

A noisy proposal span $\bm{x}^{\mathrm{Mo}}_t$ is taken as an example. To extract the corresponding feature from the video memory $\bm{M}_\mathrm{V}$, the RoI feature extract module are employed as,

\begin{equation}
    \bm{f}_\mathrm{p}=\rho[{\phi(\bm{M}_\mathrm{V},\bm{x}^{\mathrm{Mo}}_t)}]
    \label{decoder_EQ}
\end{equation}

\noindent where $\phi$ is 1-D RoI alignment and $\rho$ is average pooling. It is straightforward to implement 1-D RoI alignment: all video frame features within the timestamp are selected as the output. Then, the features $\bm{f}_\mathrm{p}$ are sent to the feature interaction module, which is structured following the guidelines in~\cite{sparsercnn}. This module includes a self-attention module, a dynamic interactive module, and a feed-forward network, designed to facilitate feature interaction among various proposals. Concurrently, timestep embeddings are integrated into the features, enabling the denoising head to know the intensity of the added noise. The moment denoising is completed by a 3-layer MLP. The prediction result $\hat{\bm{x}}^{\mathrm{Mo}}_0$ is a 2-D vector representing the temporal span's centre point and width.

Our decoder distinctly diverges from DETR, as it forgoes the utilization of learnable queries updated through backpropagation. Instead, it explicitly models the features of the proposal span via temporal RoI alignment and refines them progressively. The diffusion models allow iterative sampling during the inference process. Newly generated temporal spans serve as the proposal spans for the next stage, regressing more precise boundaries. The entire decoder is employed repeatedly, with parameter sharing across different steps.

\subsection{Highlight Detection Branch}

The highlight detection branch is designed to obtain saliency scores for video clips from the generative modelling perspective. Unlike the moment retrieval, which optimizes multiple noisy spans, the highlight detection branch produces only one saliency score vector containing all video clip saliency scores. This vector reflects the correlation between the query and the video clip. The learning objective of this generation process is equivalent to approximating the query-based clip distribution.

In specific, an attentive pooling operator is adopted to aggregate the word embedding sequence $ \bm{E}_\mathrm{Q}$ into a sentence representation $ \bm{E}_\mathrm{S}= \bm{A} \bm{E}_\mathrm{Q} \in \mathbb{R}^{1 \times D} $. $\bm{A} = \text{softmax}(\bm{W}\bm{E}_\mathrm{Q})$, and $\bm{W}$ is a learnable embedding. This sentence representation can comprehensively capture the query's content. Subsequently, an enhanced cross-attention mechanism is employed to reweight the clip representations. For the sentence embedding $\bm{E}_\mathrm{S}$, the video clip embedding $\bm{E}_\mathrm{V}$, and the diffusion step $t$,

\begin{equation}
\bm{Q}, \bm{K}, \bm{V} = \bm{W}_{\{\mathrm{Q}, \mathrm{K}, \mathrm{V}\}} [\bm{E}_{\{\mathrm{S},\mathrm{V},\mathrm{V}\}} + Emb(t)],
\end{equation}

\noindent where $\bm{W}_{\{\mathrm{Q}, \mathrm{K}, \mathrm{V}\}}$ represents three learnable projection embeddings. The function $Emb(\cdot)$ maps the diffusion step $t$ into $D$-dimensional embedding. The diffused saliency score vector $\bm{x}^{\mathrm{Sa}}_t $, along with $\bm{Q}$ and $\bm{K}$, is utilized to reweight the clip representations $\bm{V}$. This process is expressed as,

\begin{equation}
\bm{E}^t_\mathrm{V} = \text{softmax}(\bm{Q}\bm{K}^T + \bm{x}^{\mathrm{Sa}}_t) \bm{V},
\end{equation}

\noindent where $ \bm{E}^t_V $ is treated as the clip features at the $ t $-th diffusion step, incorporating information from the textual query. An MLP is used as the denoising decoder, and a linear layer is ultimately utilized to compute the output distribution.

In addition, the saliency calculation strategy from the baseline method~\cite{moment-detr} is retained and formulated as,

\begin{equation}
\bm{s}_{\text{dis}} = \text{Linear}(\bm{M}_V).
\label{saliency_discriminative}
\end{equation}

This strategy promotes interaction between textual words and video clips, optimized through a discriminative loss function. The final saliency score vector is represented as $ \bm{s} = \bm{s}_{\text{gen}} + \bm{s}_{\text{dis}} $. This can be perceived as a combination of generative and discriminative models, as well as a fusion between global and local perspectives.

\begin{figure}[!t]
  \centering
  \includegraphics[width=1\linewidth]{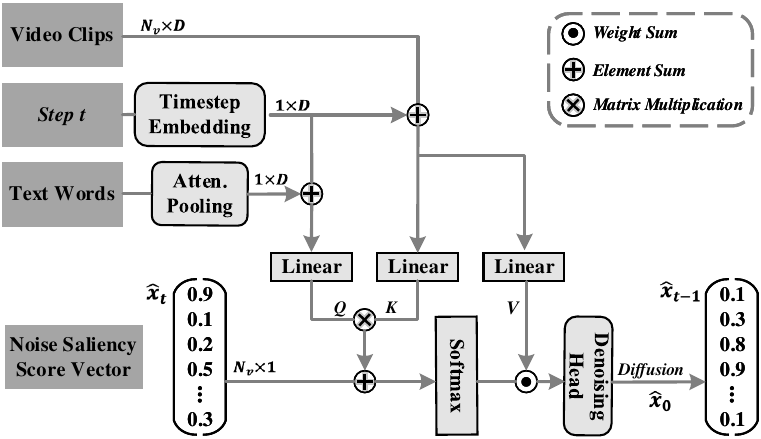}
  \caption{The pipeline of the highlight detection branch. In each sampling step, the denoising head directly predicts the distribution of $\bm{\hat{x}}_0$ based on $\bm{\hat{x}}_{t}$. For iterative refinement, $\bm{\hat{x}}_{t-1}$ can be derived from $\bm{\hat{x}}_0$ through  Eq.\ref{Diff_infence_diffusion} and serves as the input for the next step.}
  \label{fg_OverviewHL}
  \vspace{-0.1 cm}
\end{figure}

\subsection{Objective Function}

Before calculating the loss, the Hungarian algorithm~\cite{hungarianmethod} is adopted to find the optimal bipartite assignment between the moment predictions and the ground truth,

\begin{equation}
\hat{\kappa} = \text{argmin}_{\kappa} \sum\nolimits_{i}^{N} \mathcal{L}_\mathrm{match} (\bm{P}^i, \hat{\bm{P}}^{\kappa (i)}),
\end{equation}

\noindent $\hat{\bm{P}}=\{(\hat{\bm{x}}^{\mathrm{Mo}}_0, \hat{\bm{p}}_c) \}_{i=1}^N$ is a set of predictions. $\bm{P}$ is a set of ground truths. $\mathcal{L}_\mathrm{match}(\cdot)$ is the matching cost function between the ground truth and the moment prediction,

\vspace{-.6 em}

\begin{equation}
    \mathcal{L}_\mathrm{match}(\bm{P}^i, \hat{\bm{P}}^{\kappa (i)})= -\hat{\bm{P}}_c^{\kappa (i)} + \mathcal{L}_\mathrm{span} (\bm{P}_s^i, \hat{\bm{P}}_s^{\kappa (i)}) .
\end{equation}

Based on the optimal assignment $\hat{\kappa}$, the loss can be obtain by the following function,

\vspace{-.9 em}

\begin{equation}
\mathcal{L} = \lambda_\mathrm{class} \mathcal{L}_\mathrm{class} + \mathcal{L}_\mathrm{span} + \lambda_\mathrm{saliency} \mathcal{L}_\mathrm{saliency} ,
\end{equation}

\noindent where $\lambda_\text{class}, \lambda_\text{saliency}\in {\mathbb{R}}$ are the trade-off hyper-parameters. $\mathcal{L}_\text{class}$ is the cross-entropy loss used to measure whether the predicted moment is classified as foreground or background correctly. $\mathcal{L}_\text{span}$ is used to measure the discrepancy between the moment of the ground truth and the prediction, formally,

\vspace{-.6em}

\begin{equation}
\mathcal{L}_\mathrm{span} = \lambda_{L1} \mathcal{L}_{L1}(\bm{P}_s^i, \hat{\bm{P}}_s^{\hat{\kappa} (i)}) + \lambda_\mathrm{iou} \mathcal{L}_\mathrm{iou}(\bm{P}_s^i, \hat{\bm{P}}_s^{\hat{\kappa} (i)}) ,
\end{equation}

\noindent where the $\mathcal{L}_{L1} $ is $L1$ loss and $\mathcal{L}_\text{iou}$ is 1-D IoU loss. The saliency loss is composed of a hinge loss $\mathcal{L}_\text{hinge}$ and a Kullback-Leibler (KL) divergence $\mathcal{L}_\text{KL}$.

\vspace{-.4 em}

\begin{equation}
\mathcal{L}_\mathrm{saliency} = \mathcal{L}_\mathrm{hinge} + \mathcal{L}_\mathrm{KL},
\end{equation}

\noindent $ L_\text{hinge}=\text{max} (0, \Delta + S^\text{low} - S^\text{high})$ is used to optimize the discriminative saliency score $\bm{s}_{\text{dis}}$. $\Delta \in {\mathbb{R}}$ is the margin. Besides, the moment denoising process in DiffusionVMR relies on the query-related video representation, which makes the quality of features obtained by the cross-modal encoder crucial for learning. The hinge loss $\mathcal{L}_\text{hinge}$ can help to optimize the features generated by the cross-modal encoder. The Kullback-Leibler divergence loss, denoted as $\mathcal{L}_{\text{KL}} = \text{KL}(\bm{x}^{\mathrm{Sa}}_0 || \hat{\bm{x}}^{\mathrm{Sa}}_0)$, is utilized to optimize the generative saliency score $\bm{s}_{\text{gen}}$. The choice is based on the ability of KL divergence to quantify the differences between probability distributions accurately. The objective of this loss is to minimize the divergence between $\bm{x}^{\mathrm{Sa}}_0$ and $\hat{\bm{x}}^{\mathrm{Sa}}_0$.

% === IV. Transistor Class-F inv Rectifier ========================================
% =================================================================================
% \newpage
%%%%%%%%%%%%%%%%%%%%%%
% Experiments
%%%%%%%%%%%%%%%%%%%%%

\section{Experiments}

In this section, the effectiveness of the proposed method is validated on both the video moment retrieval and highlight detection tasks. The datasets, implementation details, and evaluation metrics are introduced in Section A. The proposed method is compared with the state-of-the-art approaches in Section B. The ablation studies on the QVHighlights dataset are discussed in Section C. Finally, the qualitative results are shown in Section D.

\begin{table*}[!t]
\centering
\caption{{\upshape Comparisons on the test split of the QVHighlights dataset for the joint moment retrieval and highlight detection tasks. $\dagger$ indicates the methods are unsuitable for direct comparison due to the extra usage of audio features. The results presented are obtained at S5@50, i.e., 50 proposal spans with 5 sampling steps. Due to limitations on submission quantity to the testing server, only the best results achieved are reported. The best results are highlighted in bold, while the second-best are underlined.}}
\renewcommand{\arraystretch}{1.1}
\begin{tabular}{lccccccc}
\toprule
\multicolumn{1}{c}{\multirow{2}{*}{Method}} & \multicolumn{5}{c}{MR}                   & \multicolumn{2}{c}{HD $\ge $ Very Good}     \\ \cline{2-8} 
\multicolumn{1}{c}{}                        & R1@0.5           & R1@0.7           & mAP@0.5           & mAP@0.75          & mAP@Avg.      & mAP       & HIT@1                        \\ \midrule
MCN~\cite{twostage2}                        & 11.41          & 2.72           & 24.94          & 8.22           & 10.67      & -         & -                             \\
CAL~\cite{cal}                              & 25.49          & 11.54          & 23.40          & 7.65           & 9.89       & -         & -                              \\
CLIP~\cite{clip}                            & 16.88          & 5.19           & 18.11          & 7.00           & 7.67       & 31.30     & 61.04                           \\
XML~\cite{xml}                              & 41.83          & 30.35          & 44.63          & 31.73          & 32.14      & 34.49     & 55.25                           \\
XML+~\cite{moment-detr}                     & 46.69          & 33.46          & 47.89          & 34.67          & 34.90      & 35.38     & 55.06                            \\
M-DETR~\cite{moment-detr}              & 52.89          & 33.02          & 54.82          & 29.40          & 30.73      & 35.69     & 55.60                             \\
{UMT}$^\dagger$~\cite{UMT}                  & 56.23          & 41.18          & 53.83          & 37.01          & 36.12      & 38.18     & 59.99                              \\
QD-DETR~\cite{qddetr}                       & \underline{62.40}  & \underline{44.98}  & \underline{62.52} & \underline{39.88} & \underline{39.86}        & \underline{38.94}         & \underline{62.40}          \\
MH-DETR~\cite{mhdetr}                       & 60.05          & 42.48          & 60.75          & 38.13          & 38.38      & 38.22     & 60.51          \\
UniVTG~\cite{univtg}                        & 58.86          & 40.86          & 57.60          & 35.59          & 35.47      & 38.20     & 60.96          \\
% \textbf{DiffusionVMR} (ours)                & \textbf{62.45} & \textbf{45.59} & \textbf{63.68} & \textbf{42.38} & \textbf{41.72}        & \textbf{39.53}         & \textbf{64.19} \\ \hline
% \textbf{DiffusionVMR} (ours)                & \textbf{63.55} & \textbf{47.60} & \textbf{63.85} & \textbf{43.60} & \textbf{42.38}        & \textbf{39.28}         & \textbf{63.81} \\ \hline
\textbf{DiffusionVMR}                       & \textbf{62.91} & \textbf{47.08} & \textbf{63.68} & \textbf{43.88} & \textbf{42.74}        & \textbf{39.80}         & \textbf{65.43} \\ \bottomrule
\end{tabular} 
\vspace{-0.3 cm}
\label{MR_QVHL}
\end{table*}

\begin{table}[!t]
\centering
\caption{{\upshape Comparisons on the test split of the Charades-STA dataset for video moment retrieval. Since the proposal spans are randomly initialized from Gaussian noise, the experiment is repeated 5 times at the same checkpoint, with the mean performance and standard reported. The results presented are obtained at S2@20.}}
\renewcommand{\arraystretch}{1.1}
\begin{tabular}{lccc}
\toprule
Method                & Feat.     & R1@0.5         & R1@0.7         \\ \midrule
SAP~\cite{twostage6}        & VGG       & 27.42          & 13.36          \\
SM-RL~\cite{reinforcement2} & VGG & 24.36          & 11.17          \\
TripNet~\cite{tripnet} & VGG       & 36.61          & 14.50          \\
MAN~\cite{anchor2}     & VGG       & 41.24          & 20.54          \\
2D-TAN~\cite{2D-TAN}  & VGG       & 40.94          & 22.85          \\
FVMR~\cite{fvmr}      & VGG       & 42.36          & 24.14          \\
UMT~\cite{UMT}        & VGG+Opti. & 49.35          & 26.16          \\
QD-DETR~\cite{qddetr} & VGG       & 52.77          & 31.13          \\
MH-DETR~\cite{mhdetr} & VGG       & \underline{55.47}          & \underline{32.41}          \\
\textbf{DiffusionVMR} & VGG       & $\textbf{57.74}_{\pm 0.11}$ & $\textbf{37.23}_{\pm 0.26}$ \\ \midrule
CAL~\cite{cal}        & SF+C      & 47.31          & 30.19          \\
2D-TAN~\cite{2D-TAN}  & SF+C      & 39.70          & 23.31          \\
VSLNet~\cite{vslnet}  & SF+C      & 47.31          & 30.19          \\
IVG-DCL~\cite{ivgdcl} & SF+C      & 50.24          & 32.88          \\
M-DETR~\cite{moment-detr} & SF+C      & 53.63          & 31.37          \\
QD-DETR~\cite{qddetr} & SF+C      & 57.31          & 32.55          \\
UniVTG~\cite{univtg}  & SF+C      & \underline{58.01}          & \underline{35.65}          \\
\textbf{DiffusionVMR} & SF+C      & $\textbf{59.66}_{\pm 0.15}$ & $\textbf{38.73}_{\pm 0.08}$  \\ \bottomrule
\end{tabular}
\vspace{-0.4 cm}
\label{MR_charades}
\end{table}

\begin{table}[t]
\centering
\caption{{\upshape Comparisons on the test split of the TACoS dataset for video moment retrieval. The experiment is repeated 5 times, with the mean performance and standard report. The results presented are obtained at S2@20.}}
\renewcommand{\arraystretch}{1.1}
\begin{tabular}{lcccc}
\toprule
Method                       & Feat.          & R1@0.1 & R1@0.3 & R1@0.5 \\ \midrule
MCN~\cite{twostage2}         & C3D          & 14.42  & -      & 5.58   \\
CTRL~\cite{twostage1}        & C3D          & 24.32  & 18.32  & 13.30  \\
TGN~\cite{anchor1}           & C3D          & 41.87  & 21.77  & 18.90  \\
%ACRN~\cite{acrn}             & C3D          & 24.22  & 19.52  & 14.62  \\
ACL-K~\cite{aclk}            & C3D          & 31.64  & 24.17  & 20.01  \\
CMIN~\cite{cmin}             & C3D          & 36.68  & 27.33  & 19.57  \\
SM-RL~\cite{reinforcement2}  & C3D          & 26.51  & 20.25  & 15.95  \\
SLTA~\cite{slta}             & C3D          & 23.13  & 17.07  & 11.92  \\
SAP~\cite{twostage6}               & C3D          & 31.15  & -      & 18.24  \\
TripNet~\cite{tripnet}       & C3D          & -      & 23.95  & 19.17  \\
2D-TAN~\cite{2D-TAN}         & C3D          & 47.59  & 37.29  & 25.32  \\
CSMGAN~\cite{csmgan}         & C3D          & 42.74  & 33.90  & 27.09  \\
FIAN~\cite{fian}             & C3D          & 39.55  & 33.87  & 28.58  \\
LGN~\cite{lgn}               & C3D          & 52.46  & 41.71  & 30.57  \\
SMRN~\cite{smrn}             & C3D          & 50.44  & 42.49  & 32.07  \\
VLG-Net~\cite{vlgnet}        & C3D          & \underline{57.21}  & \underline{45.46}  & \underline{34.19}  \\
\textbf{DiffusionVMR}        & C3D          & $\textbf{58.57}_{\pm 0.17}$  & $\textbf{47.41}_{\pm 0.14}$  & $\textbf{36.79}_{\pm 0.27}$  \\ \midrule
2D-TAN~\cite{2D-TAN}         & SF+C         & -      & 40.01  & 27.99  \\
VSLNet~\cite{vslnet}         & SF+C         & -      & 35.54  & 23.54  \\
M-DETR~\cite{moment-detr}    & SF+C         & -      & 37.97  & 24.67  \\
UniVTG~\cite{univtg}         & SF+C         & -      & \underline{51.44}  & \underline{34.97}  \\
\textbf{DiffusionVMR} & SF+C         & $\textbf{63.84}_{\pm 0.12}$  & $\textbf{52.90}_{\pm 0.06}$  & $\textbf{38.02}_{\pm 0.19}$  \\ \bottomrule
\end{tabular}
\vspace{-0.4 cm}
\label{MR_tacos}
\end{table}

\subsection{Experiment Setting}
\noindent \textbf{Datasets.} Experiments are conducted on the following five publicly used datasets.

\begin{itemize}[] % leftmargin=0.265cm, itemindent=0cm

\item \textbf{QVhighlights}~\cite{moment-detr} {is a benchmark designed for the joint video moment retrieval and highlight detection tasks. It consists of $10,310$ samples annotated with human-written text queries. Experiments are conducted on the standard split, i.e., $7,218$ query-moment pairs for training, $1,550$ for validation, and $1,512$ for testing. Notably, the test splits are required to be evaluated on the server, which allows QVHighlights to ensure a fair assessment.}

\item \textbf{Charades-STA}~\cite{twostage1} {is a dataset created by adding temporal annotations on Charades dataset~\cite{charadesPlus}. It comprises $16,124$ query-moment pairs tailored for video moment retrieval tasks. The average length of videos and target moments are $30.60$ and $8.09$ seconds, respectively. Following previous work~\cite{moment-detr,UMT}, 12,404 query-moment pairs are utilized for training and 3,720 for testing.}

\item \textbf{TACoS}~\cite{Tacos} {include 127 videos selected from the MPII Cooking dataset~\cite{TacosPlus}, specially used for video moment retrieval tasks. It contains $18,818$ moment-query pairs. Experiments are conducted on the standard split, i.e., $10,146$ query-moment pairs for training, $4,589$ for validation, and $4,083$ for testing.}

\item \textbf{TVSum}~\cite{tvsum}{ is a dataset used for video highlight detection tasks. It includes 50 videos across 10 domains, each annotated with 20 saliency scores. Video titles serve as text queries. The dataset splits and settings are all following the previous works~\cite{UMT,univtg}.}

\item \textbf{YouTube Highlights}~\cite{youtube}{ comprises 433 videos across 6 domains designed for highlight detection tasks. Since video titles are not provided, the domain name of each video is used as the text query. The configuration of this dataset is also in line with the previous works~\cite{UMT,univtg}.}

\end{itemize}

\noindent \textbf{Implementation Details.} PyTorch is adopted as the experimental environment. The input frames are limited to $75$ per video and the words are limited to $32$ for a query. The beyond tokens are truncated. The cross-modal encoder consists of six self-attention layers and the moment denoising decoder comprises two cascading denoising layers. The hidden dimension of the model is set to $D=256$. The proposed model is trained for $300$ epoch, and the checkpoint with the best validation performance is selected for testing. The weights of the model are randomly initialized using Xavier initialization~\cite{Xavier}. The optimizer is AdamW with the initial learning rate $1e^{-4}$ and the weight decay $1e^{-4}$. The hyperparameters used in the loss function are set to $\lambda_{L1}=5$, $\lambda_{iou}=2$, $\lambda_{class}=2$ and $\lambda_{saliency}=1$. The maximum diffusion step is set to $T=1000$. The number of proposals in training is gradually increased from $1$ to $20$. In inference, the proposal numbers and sampling steps are determined through a grid search strategy. All experiments are conducted on 1$\times$NVIDIA RTX 3090 GPU.  The following dataset features are used in this experiment: officially released SlowFast and CLIP features for the QVHighlights dataset~\cite{moment-detr}, SlowFast and CLIP features for the Charades-STA ~\cite{moment-detr}, VGG features for the Charades-STA ~\cite{UMT}, and C3D features for the TACoS ~\cite{vlgnet}, as well as SlowFast and CLIP features for TVSum and YouTube Highlights dataset~\cite{univtg}.

\smallskip
\noindent \textbf{Metrics.} Evaluation metrics for video moment retrieval tasks include Recall@1 with IoU thresholds M (R1\text{@}M), mAP with IoU thresholds M (mAP\text{@}M), and average mAP (mAP\text{@}Avg). Higher IoU values indicate more precise moment matching. Following the convention, $\text{R1@}\{0.1, 0.3, 0.5, 0.7\}$ and $\text{mAP@}\{0.5, 0.75\} $ are used in our experiments.For highlight detection in the QVhighlight dataset, the metrics of mAP and Hit@1 are employed, with the thresholds set to `Very Good'. For the TVSum and YouTube Highlights datasets, the mAP and Top-5 mAP metrics are adopted respectively~\cite{UMT,univtg}. All metrics used in the experiments are as high as better.

\begin{table*}[!t]
\centering
\caption{{\upshape Highlight Detection results of Top-5 mAP on TVSum. $\dagger$ denotes using audio modality. The experiment is repeated 5 times, with the mean performance and standard report. The sampling step is tailored for different domains.}}
\renewcommand{\arraystretch}{1.1}
\renewcommand\tabcolsep{5pt}
\begin{tabular}{lccccccccccc}
\toprule
\multicolumn{1}{c}{\textbf{Method}}     & {VT}    & {VU}   & {GA}   & {MS}   & {PK}   & {PR}   & {FM}   & {BK}   & {BT}   & {DS}   & \textbf{Avg.} \\ \midrule
sLSTM \cite{sLSTM}                        & $41.1$ & $46.2$  & $46.3$ & $47.7$ & $44.8$ & $46.1$ & $45.2$ & $40.6$ & $47.1$ & $45.5$ & $45.1$ \\ 
SG \cite{SG}                              & $42.3$ & $47.2$  & $47.5$ & $48.9$ & $45.6$ & $47.3$ & $46.4$ & $41.7$ & $48.3$ & $46.6$ & $46.2$ \\
LIM-S \cite{lessismore}                   & $55.9$ & $42.9$  & $61.2$ & $54.0$ & $60.4$ & $47.5$ & $43.2$ & $66.3$ & $69.1$ & $62.6$ & $56.3$ \\
Trailer \cite{trailer}                    & $61.3$ & $54.6$  & $65.7$ & $60.8$ & $59.1$ & $70.1$ & $58.2$ & $64.7$ & $65.6$ & $68.1$ & $62.8$ \\
SL-Module \cite{SL_module}                & $86.5$ & $68.7$  & $74.9$ & $\mathbf{86.2}$ & $79.0$ & $63.2$ & ${58.9}$ & ${72.6}$ & ${78.9}$ & $64.0$ & ${73.3}$ \\
UniVTG~\cite{univtg}                    & $83.9$ & $85.1$  & $89.0$ & $80.1$ & $\mathbf{84.6}$ & $81.4$ & $70.9$ & $91.7$ & $73.5$ & $69.3$ & $81.0$ \\ \midrule

MINI-Net$^\dagger$ \cite{miniNet}         & $80.6$ & $68.3$  & $78.2$ & $81.8$ & $78.1$ & $65.8$ & $57.8$ & $75.0$ & $80.2$ & $65.5$ & $73.2$ \\
TCG$^\dagger$ \cite{tcg}                  & $85.0$ & $71.4$  & $81.9$ & $78.6$ & $80.2$ & $75.5$ & $71.6$ & $77.3$ & $78.6$ & $68.1$ & $76.8$ \\
Joint-VA$^\dagger$ \cite{jointVA}         & $83.7$ & $57.3$  & $78.5$ & $86.1$ & $80.1$ & $69.2$ & $70.0$ & $73.0$ & $\mathbf{97.4}$ & $67.5$ & $76.3$ \\
UMT$^\dagger$\cite{UMT}                   & $87.5$ & $81.5$  & $88.2$ & $78.8$ & $81.5$ & $\mathbf{87.0}$ & $\mathbf{76.0}$ & ${86.9}$ & ${84.4}$ & $\mathbf{79.6}$ & ${83.1}$ \\\midrule

\textbf{DiffusionVMR}         & \begin{tabular}[c]{@{}c@{}}{$\mathbf{87.6}$}\\$_{\pm0.01}$\end{tabular} & \begin{tabular}[c]{@{}c@{}}{$\mathbf{93.5}$}\\$_{\pm0.02}$\end{tabular}  & \begin{tabular}[c]{@{}c@{}}{$\mathbf{94.1}$}\\$_{\pm0.02}$\end{tabular}  & \begin{tabular}[c]{@{}c@{}}{84.9}\\$_{\pm0.01}$\end{tabular}  & \begin{tabular}[c]{@{}c@{}}{79.4}\\$_{\pm0.05}$\end{tabular}  & \begin{tabular}[c]{@{}c@{}}{79.6}\\$_{\pm0.02}$\end{tabular}  & \begin{tabular}[c]{@{}c@{}}{66.8}\\$_{\pm0.04}$\end{tabular}  & \begin{tabular}[c]{@{}c@{}}{$\mathbf{92.9}$}\\$_{\pm0.02}$\end{tabular}  & \begin{tabular}[c]{@{}c@{}}{85.8}\\$_{\pm0.01}$\end{tabular}  & \begin{tabular}[c]{@{}c@{}}{$\mathbf{79.6}$}\\$_{\pm0.03}$\end{tabular}  & $\mathbf{84.4}$ \\   \bottomrule

\end{tabular}
\vspace{-0.1 cm}
\label{HL_TVSum}
\end{table*}

\subsection{Comparison with State-of-the-arts}
In this section, the proposed DiffusionVMR is compared with the state-of-the-art methods on five benchmarks.

\begin{table}[!t]
\centering
\caption{{\upshape Highlight Detection results of mAP on YouTube HL. $\dagger$ denotes using audio modality. The sampling step is tailored for different domains.}}
\renewcommand{\arraystretch}{1.1}
\renewcommand\tabcolsep{4.05pt}
\begin{tabular}{lccccccc}
\toprule
\multicolumn{1}{c}{\textbf{Method}}    & Dog     & Gym.        & Par.     & Ska.    & Ski.     & Sur.     & \textbf{Avg.} \\ \midrule
RRAE \cite{RRAE}              & 49.0    & 35.0        & 50.0     & 25.0    & 22.0     & 49.0     & 38.3 \\  
GIFs \cite{Video2gif}         & 30.8    & 33.5        & 54.0     & 55.4    & 32.8     & 54.1     & 46.4\\
LSVM~\cite{youtube}           & 60.0    & 41.0        & 61.0     & 62.0    & 36.0     & 61.0     & 53.6\\
LIM-S~\cite{lessismore}       & 57.9    & 41.7        & 67.0     & 57.8    & 48.6     & 65.1     & 56.4 \\
SL-Module~\cite{SL_module}    & 70.8    & 53.2        &77.2      & 72.5    & 66.1     & 76.2     & 69.3 \\
QD-DETR~\cite{qddetr}         & 72.2    & 77.4        & 71.0   & 72.7    & 72.8     & 80.6     & 74.4 \\
UniVTG~\cite{univtg}          & 71.8    & 76.5        & 73.9     &73.3     & 73.2     & 82.2     & 75.2 \\ \midrule
MINI-Net$^\dagger$~\cite{miniNet}    & 58.2    & 61.7        & 70.2     &72.2     & 58.7     & 65.1     & 64.4 \\
TCG$^\dagger$~\cite{tcg}      & 55.4    & 62.7        & 70.9     & 69.1    & 60.1     & 59.8     & 63.0 \\
Joint-VA$^\dagger$~\cite{jointVA}    & 64.5    & 71.9        & 80.8     & 62.0    & 73.2     & 78.3     & 71.8 \\
UMT$^\dagger$~\cite{UMT}      & 65.9    & 75.2    &\textbf{81.6}  & 71.8   & 72.3 &\textbf{82.7} & 74.9 \\ \midrule
\textbf{DiffusionVMR}         & \begin{tabular}[c]{@{}c@{}}{\textbf{74.1}}\\$_{\pm0.01}$\end{tabular} & \begin{tabular}[c]{@{}c@{}}{\textbf{78.7}}\\$_{\pm0.01}$\end{tabular}  & \begin{tabular}[c]{@{}c@{}}{74.2}\\$_{\pm0.03}$\end{tabular}  & \begin{tabular}[c]{@{}c@{}}{\textbf{73.8}}\\$_{\pm0.01}$\end{tabular}  & \begin{tabular}[c]{@{}c@{}}{\textbf{73.9}}\\$_{\pm0.02}$\end{tabular}  & \begin{tabular}[c]{@{}c@{}}{80.8}\\$_{\pm0.02}$\end{tabular}  & \textbf{75.9}  \\   \bottomrule
\end{tabular}
\vspace{-0.3 cm}
\label{HL_YouTube}
\end{table}

\begin{table*}[!t]
\centering
\caption{{\upshape Effectiveness of different sampling steps on QVHighlights val split. The results presented are obtained at S@20.}}
\renewcommand{\arraystretch}{1.2}
\begin{tabular}{ccccccc}
\hline
\multirow{2}{*}{Sampling Step} & \multicolumn{3}{c}{MR}                   & \multicolumn{2}{c}{HD $\ge $ Very Good}                                      & \multirow{2}{*}{Inference Time (s)} \\ \cline{2-6} 
                               & R1@0.5                 & R1@0.7                  & mAP@Avg                     & mAP                         & HIT@1                         \\ \hline
1                              & $62.47_{\pm 0.47}$     & $46.92_{\pm 0.24}$      & $41.93_{\pm 0.36}$       & $39.02_{\pm 0.20}$          & $62.15_{\pm 0.23}$   & \textbf{11.875}          \\
2                              & $63.95_{\pm 0.27}$     & $48.78_{\pm 0.18}$      & $43.06_{\pm 0.11}$       & $39.49_{\pm 0.14}$          & $62.68_{\pm 0.83}$    & 20.068       \\
3                              & $63.93_{\pm 0.25}$     & $48.83_{\pm 0.17}$      & $43.19_{\pm 0.20}$       & $39.58_{\pm 0.10}$          & $62.95_{\pm 0.28}$    & 31.184      \\
4                              & $\textbf{64.05}_{\pm 0.28}$     & $49.34_{\pm 0.26}$      & $43.48_{\pm 0.09}$       & $39.60_{\pm 0.03}$          & $63.17_{\pm 0.26}$    & 39.073     \\
5                              & $63.97_{\pm 0.20}$     & $\textbf{49.52}_{\pm 0.29}$      & $\textbf{43.72}_{\pm 0.03}$          & $39.61_{\pm 0.06}$           & $63.16_{\pm 0.07}$     & 49.581     \\
6                              & $63.90_{\pm 0.21}$     & $49.32_{\pm 0.41}$      & $43.52_{\pm 0.12}$       & $\textbf{39.70}_{\pm 0.05}$          & $\textbf{63.67}_{\pm 0.19}$    & 58.508          \\ \hline
\end{tabular}
\label{tb_samplingstep}
% \vspace{-0.1 cm}
\end{table*}

\smallskip
\noindent\textbf{Joint Moment Retrieval and Highlight Detection.} The QVHighlights dataset offers official features and online test evaluation, facilitating the generation of fair and persuasive results. Table~\ref{MR_QVHL} shows the performance comparison on the QVHighlights test split for joint moment retrieval and highlight detection tasks. Due to limitations on submission quantity to the testing server, only the best results achieved are reported. As observed, DiffusionVMR achieves the best performance across all metrics with a clear margin. Compared to the baseline Moment-DETR, DiffusionVMR achieves comprehensive improvement, as reflected by the significant increase in the mAP. It demonstrates the effectiveness of the proposed approach. In addition, DiffusionVMR is compared with several advanced methods. Specifically, in moment retrieval tasks, our method achieves a 2.88 increase in average mAP compared to the state-of-the-art QD-DETR, exhibiting notable gains of +4 in more stringent mAP@0.75 metrics. Conversely, QD-DETR excels only at lower IoU thresholds, such as R1@0.5 and mAP@0.5. For highlight detection, DiffusionVMR gains +0.86 in mAP and +3.03 in HIT@1 compared to QD-DETR. This superiority of DiffusionVMR is attributed to the diffusion model's ability to refine boundaries iteratively. Although UMT additionally incorporates audio features to enhance cross-modal interaction, its performance still lags behind that of DiffusionVMR, which uses only visual features.

\smallskip
\noindent\textbf{Moment Retrieval.} In this subsection, the proposed DiffusionVMR is compared with a series of mainstream moment retrieval methods on two widely used benchmarks. It should be noted that the proposal spans in inference are initially generated from Gaussian noise, resulting in slight variations in the outcomes due to different initial values. Consequently, the experiment is conducted five times consecutively at the same checkpoint, with both the mean performance and standard deviation reported. Tables~\ref{MR_charades} and~\ref{MR_tacos} show the performance comparison for the Charades-STA and TACoS test split, respectively. The features used in this subsection are enumerated in the tables to ensure fair comparisons. 

A detailed analysis of both tables reveals three critical insights. Firstly, the effectiveness of the DiffusionVMR is further substantiated. In the setting of using convolution features, DiffusionVMR achieves marked enhancements in performance compared to both proposal-based approaches (e.g., 2D-TAN) and proposal-free methods (e.g., QD-DETR). This trend persists even with SlowFast$+$CLIP features, where DiffusionVMR substantially outperforms the baseline Moment-DETR. Secondly, DiffusionVMR exhibits strong generalizability. The proposed method achieves consistent improvements across various features and datasets, indicating its applicability to diverse datasets with distinct characteristics. Thirdly, DiffusionVMR shows more significante improvements in stricter metrics, which is align with our motivation. This demonstrates the effectiveness of integrating the diffusion models. Overall, these findings highlight the effectiveness and generalizability of our method in moment retrieval tasks.

\smallskip
\noindent\textbf{Highlight Detection.} The proposed DiffusionVMR is also evaluated on Highlight Detection benchmarks, including YouTubeHL and TVSum. These benchmarks encompass a range of domains, with each domain considered as an individual split. The model's parameters and diffusion steps are adjusted to align with these domains` specific needs. The results are shown in Tables~\ref{HL_TVSum} and~\ref{HL_YouTube}. Overall, DiffusionVMR outperforms all competing methods in terms of average mAP metrics, regardless of whether they incorporate audio features. In the case of the YouTubeHL dataset, DiffusionVMR consistently achieves improvements across its various domains. In contrast, its performance on the TVSum dataset exhibits notable variation among domains (e.g., VU versus FM). This phenomenon can be ascribed to the dataset's limited scale.

\begin{figure}[!t]
  \centering
  \includegraphics[width=0.98\linewidth]{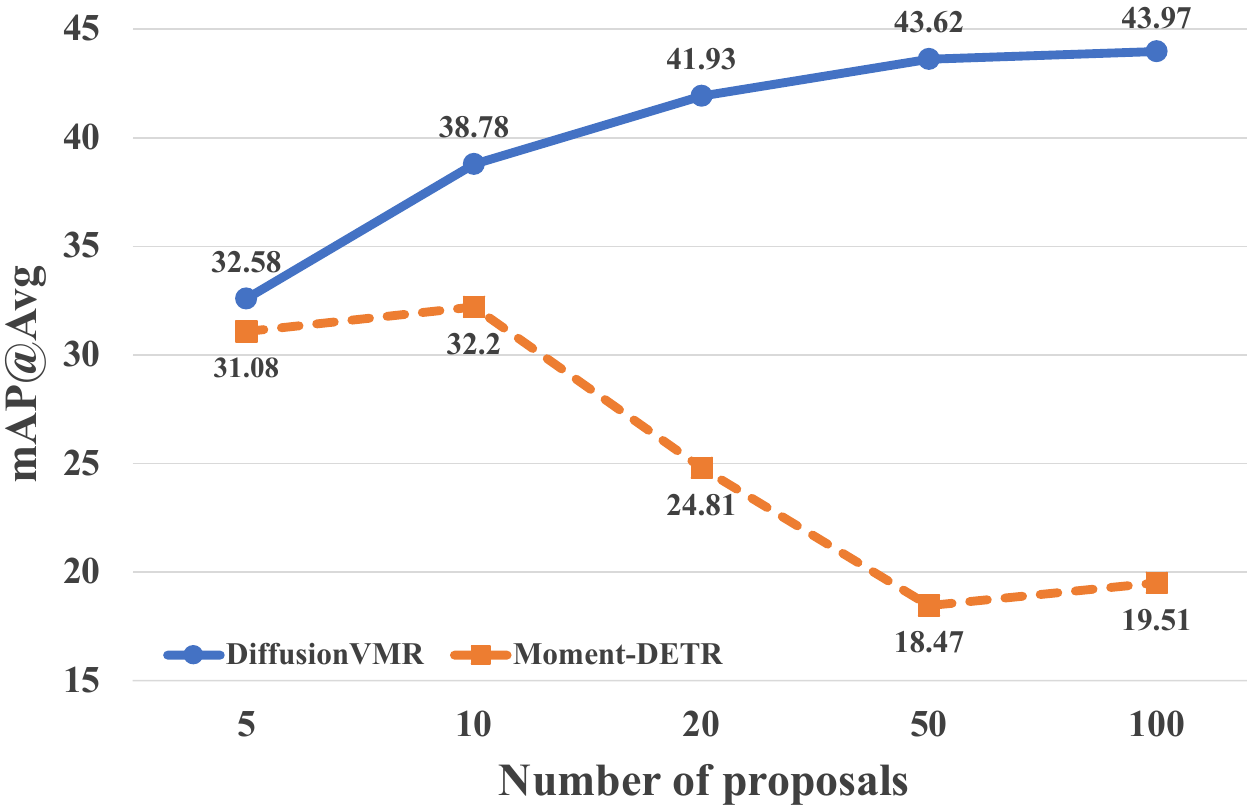}
  \caption{Effectiveness of different numbers of proposal spans on QVHighlights val split. The training and inference of DiffusionVMR are decoupled. As the proposal quantity increases during inference, the model performance steadily improves.}
  \label{fg_proposalspan}
  \vspace{-0.3 cm}
\end{figure}

\subsection{Ablation Studies}

\smallskip
\noindent\textbf{Number of Proposal Spans.} Decoupled training and inference is a crucial characteristic of DiffusionVMR. It allows the use of an arbitrary number of proposal spans for moment retrieval during inference without needing consistency with the training phase. This section investigates how the proposal quantity affects model performance in inference. The proposal number is set in the range of $5$ to $100$ and the results are reported in terms of the mAP\text{@}Avg metric. As a comparison, the performance variation of Moment-DETR under different moment queries is also reported. Figure~\ref{fg_proposalspan} shows that the performance of DiffusionVMR steadily improves as the proposal quantity increases. More random proposals provide diversified initialization for retrieval. In contrast, Moment-DETR shows performance degradation when the number of moment queries exceeds $10$. It makes sense to decouple training and inference. DiffusionVMR can be flexibly applied to tasks with different numbers of targets without re-training. Additionally, it eliminates the need to maintain a mass of proposal spans as the support set for retrieval.

\begin{figure}[!t]
  \centering
  \includegraphics[width=0.98\linewidth]{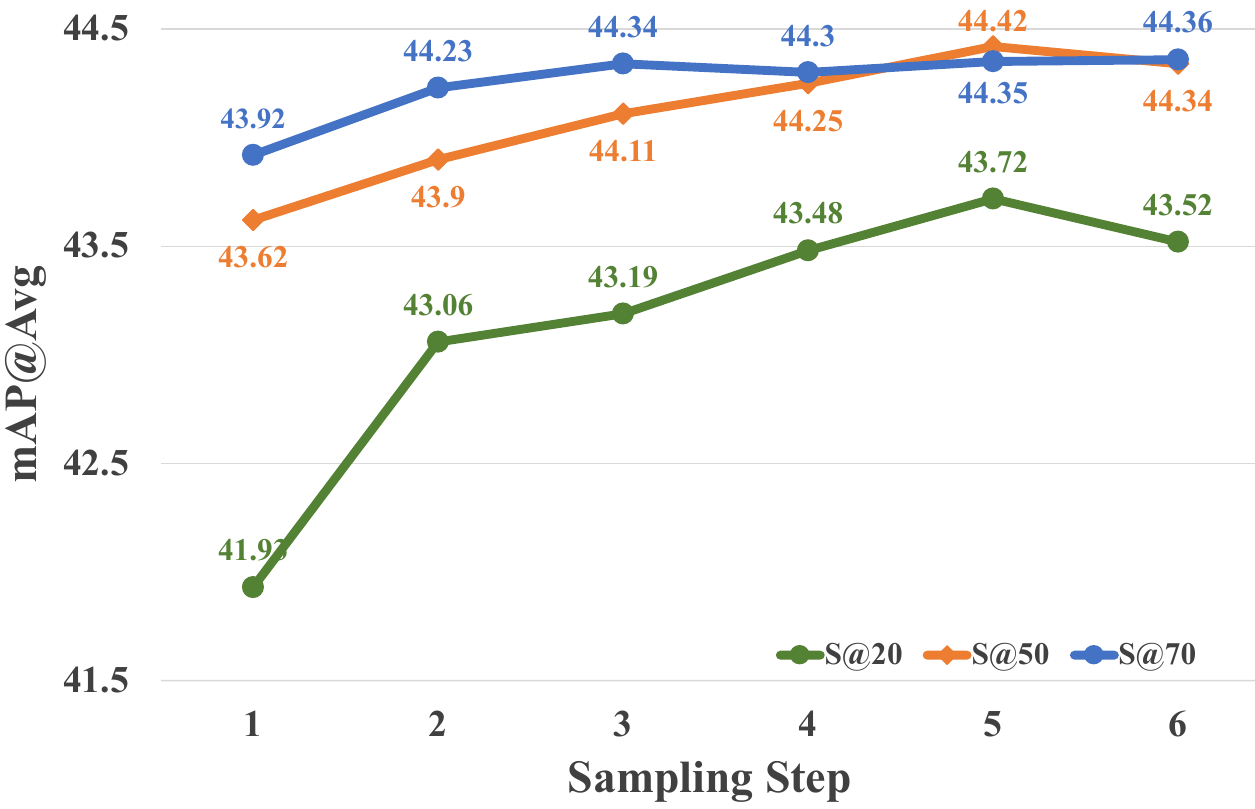}
  \caption{Effectiveness of different sampling steps on QVHighlights val split. S@20 indicates that DiffusionVMR is evaluated under 20 proposal spans using different sampling steps. For all cases, the accuracy increases with refinement times.}
  \label{fg_samplingstep}
  \vspace{-0.3 cm}
\end{figure}

\smallskip
\noindent\textbf{Progressive Refinement.} 
Another distinctive characteristic of DiffusionVMR is the capability to iteratively refine target boundaries by varying the sampling steps during inference. This section explores the effects of progressive refinement by increasing the sampling steps from $1$ to $6$. The results are detailed in Table~\ref{tb_samplingstep} and Figure~\ref{fg_samplingstep}. 

\begin{figure}[t]
  \centering
  \includegraphics[width=\linewidth]{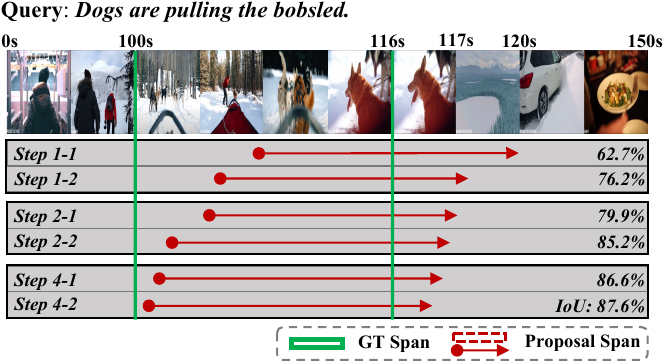}
  \caption{Effectiveness across different denoising layer on QVHighlights val split.}
  \label{fg_denoising_layer}
  \vspace{-0.3 cm}
\end{figure}

Several insights can be drawn from the results. Firstly, all metrics exhibit an increase alongside the growing number of iterations in both moment retrieval and highlight detection tasks. It demonstrates that the model inherits the advantages of diffusion models. Secondly, performance enhancements are more pronounced under stricter metrics, confirming the motivation behind our proposed method. Thirdly, there is an upper limit to model performance, while more number of iterations increases the time cost. Lastly, models with fewer proposal spans benefit substantially from the iterative refinement in moment retrieval tasks. For example, the average mAP of a model with 20 proposal spans improves from $41.93$ ($1$ step) to $43.72$ ($5$ steps). Additionally, refinement in moment retrieval tasks also transpires across different denoising layers of the Moment Denoising Decoder. As depicted in Figure~\ref{fg_denoising_layer}, the IoU metric improves with advancing layer numbers, signifying more precise proposal span boundaries.

\begin{table*}[!t]
\centering
\renewcommand{\arraystretch}{1.1}
\caption{{\upshape Effectiveness of the different model components on QVHighlights val split.}}
\begin{tabular}{cc|ccccc}
\hline
\multirow{2}{*}{Row}            &\multirow{2}{*}{Method}           & \multicolumn{3}{c}{MR}                           & \multicolumn{2}{c}{HD $\ge $ Very Good} \\ \cline{3-7} 
                                &                                  & R1@0.5              & R1@0.7               & mAP@Avg                    & mAP             & HIT@1         \\ \hline
1                               & \multicolumn{1}{l|}{DiffusionVMR} & $\textbf{64.80}_{\pm 0.57}$  & $\textbf{49.59}_{\pm 0.50}$  & $\textbf{43.60}_{\pm 0.16}$        & $\textbf{39.80}_{\pm 0.09}$      & $\textbf{63.97}_{\pm 0.40}$   \\
2                               & \multicolumn{1}{l|}{$-$ Saliency Diffusion Training}  & $64.38_{\pm 0.36}$  & $48.74_{\pm 0.14}$   & $43.34_{\pm 0.12}$        & $39.08_{\pm 0.00}$      & $62.61_{\pm 0.00}$    \\
3                               & \multicolumn{1}{l|}{$-$ Saliency Denoising Branch}    & $62.26_{\pm 0.26}$  & $46.26_{\pm 0.12}$   & $41.75_{\pm 0.10}$         & $38.05_{\pm 0.00}$    & $60.65_{\pm 0.00}$     \\
4                               & \multicolumn{1}{l|}{$-$ Moment Diffusion Training}    & $60.13_{\pm 0.00}$  & $42.65_{\pm 0.00}$   &           $38.95_{\pm 0.00}$          & $37.74_{\pm 0.00}$       & $58.76_{\pm 0.00}$          \\
% 5                               & \multicolumn{1}{l|}{$-$ Dynamic Conv.}                & $55.61_{\pm 0.00}$   & $38.45_{\pm 0.00}$    & $35.12_{\pm 0.00}$          & $36.03_{\pm 0.00}$       & $57.48_{\pm 0.00}$         \\
5                               & \multicolumn{1}{l|}{$-$ Dynamic Conv.}                & $56.39_{\pm 0.00}$   & $39.16_{\pm 0.00}$    & $34.51_{\pm 0.00}$          & $36.76_{\pm 0.00}$       & $58.13_{\pm 0.00}$         \\
% 6                               & \multicolumn{1}{l|}{$-$ 4 Encoder}                    & $55.42_{\pm 0.00}$   & $36.52_{\pm 0.00}$    & $32.25_{\pm 0.00}$          & $35.96_{\pm 0.00}$       & $55.87_{\pm 0.00}$         \\ \hline
6                               & \multicolumn{1}{l|}{$-$ 4 Encoder}                    & $55.42_{\pm 0.00}$   & $36.52_{\pm 0.00}$    & $32.25_{\pm 0.00}$          & $36.03_{\pm 0.00}$       & $57.48_{\pm 0.00}$         \\ \hline
\end{tabular}
\label{tb_moduleablation}
\vspace{-0.1 cm}
\end{table*}

\smallskip
\noindent\textbf{Effectiveness of the Module Components.} In this section, each component is sequentially dropped from the DiffusionVMR to evaluate the effectiveness of the proposed framework. The results are shown in Table~\ref{tb_moduleablation}. Overall, each component brings improvements. Specifically, the saliency diffusion training strategy is removed in Row 2. This alteration led the saliency score $S_{gen}$ to degrade to a mere cosine similarity between clip embeddings and sentence representations. In Row 3, the highlight detection branch is completely abandoned. The saliency score obtained in this setting is solely based on Eq.~\ref{saliency_discriminative}. The results emphasize the pivotal role of the branch in improving highlight detection tasks and its value in moment retrieval due to its restriction on the underlying features. The moment diffusion training strategy is dropped in Row 4, which causes the model to fail to generate the target result from the noise. As compensation, an equivalent number of learnable queries is introduced following previous work~\cite{sparsercnn}. This change leads to a significant decline in almost all metrics, confirming the efficacy of the proposed moment diffusion training strategy. In the setting of Rows 5 and 6, the dynamic convolution and the additional four layers of the cross-modal encoder are removed, respectively. The final model architecture is degraded to be consistent with the baseline Moment-DETR.

\begin{table}[!t]
\centering
\caption{{\upshape Effectiveness of different components of the loss on QVHighlights val split.}}
\renewcommand{\arraystretch}{1.1}
\setlength{\tabcolsep}{1.1mm}{
\begin{tabular}{cccccccc}
\hline
\multicolumn{2}{c}{$\mathcal{L}_\mathrm{Span}$}    & \multicolumn{2}{c}{$\mathcal{L}_\mathrm{Saliency}$}   & \multirow{2}{*}{$\mathcal{L}_{cls.}$}    & \multicolumn{2}{c}{MR} & \multicolumn{1}{c}{HD}   \\ \cline{1-4} \cline{6-8}
$\mathcal{L}_1$   & $\mathcal{L}_{iou}$   & $\mathcal{L}_\mathrm{hinge}$   & $\mathcal{L}_\mathrm{KL}$   &                                    & R1@0.7              & mAP@Avg             & mAP       \\ \hline
                  & \checkmark            & \checkmark              & \checkmark               & \checkmark              & $47.91_{\pm 0.20}$  & $43.07_{\pm 0.13}$  & $39.49_{\pm 0.09}$   \\
\checkmark        &                       & \checkmark              & \checkmark               & \checkmark              & $41.09_{\pm 0.30}$  & $34.84_{\pm 0.12}$  & $38.90_{\pm 0.10}$   \\
\checkmark        & \checkmark            &                         & \checkmark               & \checkmark              & $48.47_{\pm 0.43}$  & $42.85_{\pm 0.13}$  & $37.09_{\pm 0.15}$   \\
\checkmark        & \checkmark            & \checkmark              &                          & \checkmark              & $46.26_{\pm 0.12}$  & $41.75_{\pm 0.10}$  & $38.05_{\pm 0.00}$   \\
\checkmark        & \checkmark            & \checkmark              & \checkmark               &                         & $34.86_{\pm 0.17}$  & $28.31_{\pm 0.08}$  & $38.89_{\pm 0.05}$   \\
\checkmark        & \checkmark            & \checkmark              & \checkmark               & \checkmark              & $\textbf{49.60}_{\pm 0.29}$  & $\textbf{43.66}_{\pm 0.11}$  & $\textbf{39.85}_{\pm 0.06}$    \\ \hline
\end{tabular}}
\label{tb_loss}
\vspace{-0.2 cm}
\end{table}

\smallskip
\noindent\textbf{Effectiveness of the Loss Components.} This section investigates the impact of various losses used in our model. The influence of each loss is evaluated by turning off one at a time. The results are reported in Table~\ref{tb_loss}. $\mathcal{L}_{span}$ is employed to constrain the accuracy of the predicted span. Deactivating either $\mathcal{L}_{1}$ or $\mathcal{L}_{iou}$ loss results in a performance decline, especially the latter having a more pronounced effect. The $\mathcal{L}_{saliency}$ is designed to optimize the saliency scores produced by the model. The omission of any saliency loss functions is observed to adversely impact the performance in highlight detection tasks. Additionally, they also play a role in moment retrieval tasks. The $\mathcal{L}_{KL}$ imposes constraints on feature representations of both video and text, while $\mathcal{L}_{hinge}$ restricts interactions within the cross-modal encoder, influencing overall model performance. The video moment retrieval task has only start and end timestamps as positive samples, which results in a data imbalance problem compared to mass frames. To address this, $\mathcal{L}_{class}$ is used to distinguish between foreground ground-truth spans and background padding, improving the model's performance on moment retrieval tasks.

\smallskip
\noindent\textbf{Effect of Matching Strategy.} In moment retrieval tasks, better results can be achieved by increasing the number of proposal spans during inference. However, if the quantity of training proposal spans substantially exceeds the ground truth count, it may lead to instability in bipartite graph matching during training. This instability stems from inconsistent optimization goals, especially during the early training phases. Therefore, it is recommended to incrementally increase the number of training proposal spans, beginning with $1$. The results shown in Table~\ref{training_strategy} corroborate that this training strategy significantly improves performance.

\begin{table}[h]
\centering
\caption{{\upshape Effectiveness of different training strategy on QVHighlights val split.}}
\renewcommand{\arraystretch}{1.1}
\begin{tabular}{cccc}
\hline
\multirow{2}{*}{Training Strategy} & \multicolumn{2}{c}{MR}  & HD   \\ \cline{2-4} 
                                   & R1@0.7     & mAP@AVG    & mAP  \\ \hline
Stable                             & $46.06_{\pm 0.17}$               & $41.37_{\pm 0.09}$               & $39.37_{\pm 0.10}$ \\
Progress                           & $\textbf{49.87}_{\pm 0.35}$      & $\textbf{43.84}_{\pm 0.08}$      & $\textbf{39.85}_{\pm 0.06}$ \\ \hline
\end{tabular}
\label{training_strategy}
\end{table}

\begin{figure}[t]
  \centering
  \includegraphics[width=\linewidth]{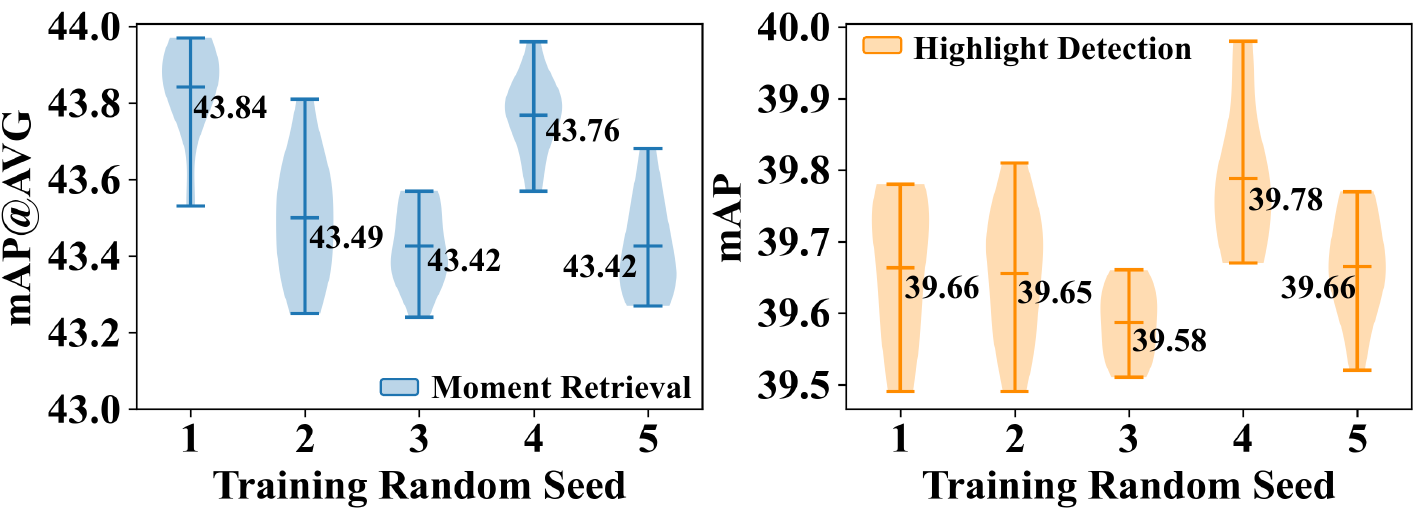}
  \caption{Statistical results over 5 independent training instances with different random seeds. Each instance is evaluated 10 times at the same checkpoint.}
  \label{fg_randomseed}
  \vspace{-0.3 cm}
\end{figure}

\smallskip
\noindent\textbf{Random Seed.} Given that the time spans and saliency scores in inference are randomly initialized from Gaussian noise, it is pertinent to consider the performance variability due to different random seeds. The robustness of DiffusionVMR is assessed by independently training five model instances, each with a unique random seed. Subsequently, each instance is evaluated ten times at the same checkpoint to gauge performance variability. Figure~\ref{fg_randomseed} shows that the average performance of the five independent trained instances in moment retrieval tasks consistently exceeded $43.4$ mAP. In highlight detection tasks, the performance hovered around $39.6$ mAP. Despite some variations in performance across instances of the model, these differences are marginal. It indicates that DiffusionVMR possesses considerable robustness and can lead to reliable performance enhancement.

\begin{figure*}[!t]
  \centering
  \includegraphics[width=1\textwidth]{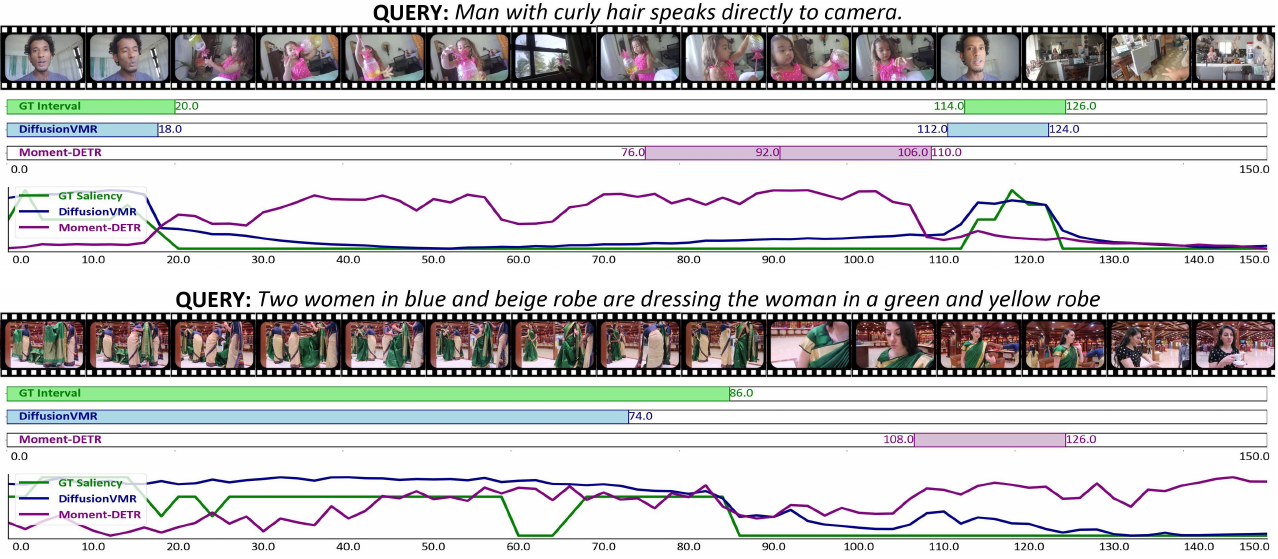}
  \caption{Visualization of joint moment retrieval and highlight detection on QVHighlights dataset.}
  \label{fg_visualization}
  \vspace{-0.5 cm}
\end{figure*}

\subsection{Visualization}

The qualitative results of DiffusionVMR are illustrated in Figure~\ref{fg_visualization}. The blue intervals and curves indicate the time spans and saliency scores predicted by our method, while the red results correspond to the Moment-DETR for comparison. As shown in the figure, DiffusionVMR not only outperforms the baseline in accurately predicting time spans but also excels in highlight detection tasks. The first example illustrates that our method can generate accurate predictions even in the video presence of multiple targets. In the second example, our model successfully identifies query-related parts in similar scenarios, showcasing its capability to tackle fine-grained challenges. Overall, DiffusionVMR exhibits a solid ability to comprehend query information and generate precise predictions.

% === IV. Conclusion ========================================
% =================================================================================

%%%%%%%%%%%%%%%%%%%%%%
% Conclusion
%%%%%%%%%%%%%%%%%%%%%

\section{Conclusion}

This paper introduces DiffusionVMR, a novel framework designed for joint video moment retrieval and highlight detection tasks. By incorporating diffusion models, DiffusionVMR reformulates the two tasks as a unified conditional denoising generation problem. For the moment retrieval task, DiffusionVMR follows the proposal-free paradigm, which initializes candidates directly from noise and progressively refines them to meaningful output. Similarly, the video highlight detection task is incorporated with the diffusion models, enabling the generation of saliency scores for each video clip from a denoising generation perspective. This denoising generation approach significantly improves the model's ability to learn text-video clip correspondences. Leveraging the iterative refinement strengths of diffusion models, DiffusionVMR achieves superior performance on stringent metrics. Moreover, decoupling the training and inference phases enhances the framework's flexibility. Experimental results on five datasets across three different settings affirm the effectiveness and flexibility of the proposed DiffusionVMR. We hope this work can inspire future research to explore the potential of the denoising generation strategy in this task.

% \section*{Acknowledgment} 

% or


% ============================================
%\appendices
%Appendix one text goes here %\cite{Roberg2010}.

%\section{}

%\section*{Acknowledgment}

%The authors would like to thank D. Root for the loan of the SWAP. The SWAP that can ONLY be usefull in Boulder...

\ifCLASSOPTIONcaptionsoff
  \newpage
\fi



%\begin{thebibliography}{1}


\bibliographystyle{IEEEtran}
\bibliography{IEEEabrv,Bibliography}

\vfill


\end{document}